\documentclass[conference,onecolumn]{IEEEtran} %
\IEEEoverridecommandlockouts
\usepackage{cite}
\usepackage{amsmath,amssymb,amsfonts}
\usepackage{algorithm}
\usepackage[binary-units]{siunitx}
\usepackage{graphicx}
\usepackage{textcomp}
\usepackage{xcolor}
\usepackage{hyperref}
\usepackage{todonotes}
\usepackage{algpseudocodex}
\usepackage{booktabs}
\usepackage{multirow}
\usepackage{graphicx}

\newcommand{\bfu}{\mathbf{u}}
\newcommand{\bfr}{\mathbf{r}}
\newcommand{\bfs}{\mathbf{s}}

\newcommand{\bfc}{\mathbf{c}}
\newcommand{\bfx}{\mathbf{x}}
\newcommand{\bfy}{\mathbf{y}}
\newcommand{\bfz}{\mathbf{z}}

\newcommand{\bfth}{\boldsymbol{\vartheta}}

\newcommand{\bfW}{\ensuremath{\mathbf{W}}}

\newcommand{\bfb}{\ensuremath{\mathbf{b}}}

\newcommand{\dti}{^{\langle t \rangle}}
\newcommand{\dtim}[1]{^{\langle t-{#1} \rangle}}

\newcommand{\hp}{{\,\mathrel{\vcenter{\hbox{\tiny$\odot$}}}\,}}

\newcommand{\bwrap}[1]{\left( #1 \right)}
\newcommand{\concat}[1]{\left[ #1 \right]}
\newcommand{\fun}[2]{{#1}\bwrap{#2}}

\def\BibTeX{{\rm B\kern-.05em{\sc i\kern-.025em b}\kern-.08em
    T\kern-.1667em\lower.7ex\hbox{E}\kern-.125emX}}
\begin{document}

\title{Language Modeling on a SpiNNaker2 Neuromorphic Chip\\
}

\author{
\IEEEauthorblockN{Khaleelulla Khan Nazeer$^{1,+}$, Mark Sch\"one$^{1}$, Rishav Mukherji$^{2}$,
\\Bernhard Vogginger$^{1}$, Christian Mayr$^{1,3}$, David Kappel$^{4}$ and Anand Subramoney$^{5}$}
\IEEEauthorblockA{\textit{$^{1}$Chair of Highly-Parallel VLSI-Systems and Neuro-Microelectronics, Technische Universität Dresden, Germany}}
\IEEEauthorblockA{\textit{$^{2}$\textit{Birla Institute of Technology and Science}, \textit{Pilani -- Goa Campus}, Goa, India }}
\IEEEauthorblockA{\textit{$^{3}$Centre for Tactile Internet (CeTI) with Human-in-the-Loop, Technische Universität Dresden, Germany}}
\IEEEauthorblockA{\textit{$^{4}$\textit{Institut f\"ur Neuroinformatik},\textit{Ruhr Universit\"at Bochum},  Bochum, Germany }}
\IEEEauthorblockA{\textit{$^{5}$\textit{Dept. of Computer Science}, \textit{Royal Holloway, University of London},
Egham, United Kingdom }}
\IEEEauthorblockA{$^+$Email:khaleelulla.khan@tu-dresden.de}
}

\maketitle

\begin{abstract}

As large language models continue to scale in size rapidly, so too does the computational power required to run them.
Event-based networks on neuromorphic devices offer a potential way to  reduce energy consumption for inference significantly.
However, to date, most event-based networks that can run on neuromorphic hardware, including spiking neural networks (SNNs), have not  achieved task performance even on par with LSTM models for language modeling. 
As a result, language modeling on neuromorphic devices has seemed a distant prospect.
In this work, we demonstrate the first-ever implementation of a language model on a neuromorphic device -- specifically the SpiNNaker2 chip -- based on a recently published event-based architecture called the EGRU.
SpiNNaker2 is a many-core neuromorphic chip designed for large-scale asynchronous processing, and the EGRU is architected to leverage such hardware efficiently while maintaining competitive task performance.
This implementation marks the first time a neuromorphic language model matches LSTMs, setting the stage for taking task performance to the level of large language models.
We also demonstrate results on a gesture recognition task based on inputs from a DVS camera.
Overall, our results showcase the feasibility of this neuro-inspired neural network in hardware, highlighting significant gains versus conventional hardware in energy efficiency for the common use case of single batch inference.

\end{abstract}


\section{Introduction}
Most deep learning systems, from edge to cloud, rely on highly regular SIMD processing.
The tremendous success of this processing paradigm has encouraged further convergence of hardware accelerators and algorithms to high throughput SIMD systems \cite{Barham2019}.
At the same time,
deep learning algorithms exhibit a surprisingly high degree of inherent sparsity, which SIMD accelerators are unable to exploit.
Experimental and theoretical studies have shown that a large fraction of connections can be removed entirely without sacrificing learning accuracy \cite{frankle2018lottery, Hoefler2021}.
Furthermore, deep learning models can operate on highly sparse representations without sacrificing precision \cite{Neil2017, li2023, subramoney2023efficient, relu}.
These findings challenge the design principles of today's SIMD-based deep learning systems from an energy efficiency perspective.
Communication is the central energy and latency cost factor in contemporary computer architectures~\cite{Horowitz2014}.
Dense matrix operations are omnipresent in deep learning and require $\mathcal{O}(n^2)$ messages for $n$-dimensional representations.
This unfavorable behavior is under growing pressure as the annual growth rate of density of computational operations in hardware is about twice as fast as the growth rate of memory and interconnect bandwidth.

In this work, we present an implementation of a sparsely connected and sparsely activated architecture implemented on an processor that can take advantage of this unstructured sparsity for energy efficiency.
More specifically, we present an implementation of a sparse network based on the EGRU~\cite{subramoney2023efficient} architecture on a SpiNNaker2 chip.
The EGRU~\cite{subramoney2023efficient} is a recently proposed event-based network that naturally exhibits high levels of activity sparsity and was shown to have high levels of task performance on language modeling and gesture recognition tasks among others.
To make full use of its potential, we implement it on the SpiNNaker2 chip, which is a digital neuromorphic system optimized for sparse communication and event-based processing \cite{gonzalez2023spinnaker}.
Our implementation operates on unstructured sparsely connected units that communicate sparse in time.
Both operations can be accelerated on SpiNNaker2, but not on conventional SIMD architectures.
We choose language modeling as our demonstrator.
Since the EGRU is a recurrent network, it is able to exploit the temporal inductive bias of sequence modeling tasks such as language modeling for computationally efficient processing.
While transformers~\cite{Vaswani2017} are the dominant architecture for language modeling, they are computationally very expensive, which makes it even more urgent to find an energy efficient alternative.
This first demonstration of the energy gains achievable using a recurrent architecture on neuromorphic hardware will set the stage for neuromorphic language modeling using even more powerful recurrent architectures~\cite{Gu2022, Peng2023}.

\section{Related Work}
Recent advances in machine learning have led to increased interest in energy-efficient hardware accelerators.  
Hardware-software co-design for machine learning accelerators have been used to target scaling to extremely large models\cite{freund2020graphcore,lie2023cerebras}.
More recently, there has been an increased focus on making transformer-based neural networks more efficient using accelerators for conventional hardware (see \cite{kim2023full} for a review). A 4-bit quantized accelerator in 5\,nm presented recently~\cite{keller202395} demonstrated high energy efficiency and throughput. 
Spiking variants of popular transformer architectures have also recently been introduced
\cite{zhang2022spike, zhu2023spikegpt, bal2023spikingbert}, but no advantage on custom hardware has been reported yet. 
Neuromorphic LSTM accelerators have been developed using FPGAs \cite{sun2018fpga}, systolic arrays \cite{conti2018chipmunk}, and memristors \cite{smagulova2019survey}. 
A hybrid LSTM/spiking neuron architecture was implemented on Intel's Loihi chip, demonstrating energy gains~\cite{rao2022long}. 
None of these LSTM-based approaches have been scaled to standard NLP benchmark tasks yet.
Spiking LSTM~\cite{Ali2020} and  EGRU~\cite{subramoney2023efficient} are two attempts at bringing event-based properties to the respective base architectures and allow for full precision gates and graded spike communication between units.
Other related approaches include Sigma-delta quantised networks that communicate only quantised changes in activations to the next layer in a feed-forward network \cite{SigmaDelta2017} and its extension to recurrent networks~\cite{Neil2017}.
An FPGA accelerator for sparsely connected and sparsely communicating Delta Networks greatly reducing required memory access, was presented in \cite{Spartus2022}.

\section{Background}
\subsection{The SpiNNaker2 System}
SpiNNaker2 is an accelerator for large-scale event-based and asynchronous processing \cite{gonzalez2023spinnaker}.
The chip consists of 152 processing elements (PEs) connected via a network-on-chip (NoC).
Each processing element is composed of an Arm M4f core, \SI{128}{\kilo\byte} SRAM, and a set of accelerators for exponential functions, random number generation and multiply-accumulate (MAC) operations.
The total of \SI{19}{\mega\byte} on-chip SRAM is accompanied by \SI{2}{\giga\byte} LPDDR4 memory.
Communication between the PEs in a single chip can be implemented by direct memory access (DMA) to other PEs' local memory.
A DMA units in every processing element enables bulk transfer without blocking the processors.
The processors receive interrupts when memory is written to a specified location of its local memory, or when their DMA instruction finished.
This allows for fully event-based implementations of sparsely communicating and asynchronously operating neural networks.
The local SRAM is organized into 4 memory banks of \SI{32}{\kilo\byte} each.
One is usually reserved for program memory, and three banks for values such as RNN weights and intermediate variables.
See Table~\ref{tab:csr-mem-footprint} for details of memory footprint of EGRU Language model.

\subsection{Event-based Gated Recurrent Unit}
The Gated Recurrent Unit (GRU) is an effective recurrent neural network that has been widely adopted for sequence modeling \cite{Cho2014}.
To reduce the communication between logical neurons, \cite{subramoney2023efficient} apply a biologically inspired thresholding mechanism to the GRU.
In this model, called Event-based Gated Recurrent Unit (EGRU), a layer consists of $n$ neurons with output $\bfy$ and state $\bfc$.
A sparse output $\bfy = (y_1, \dots, y_n)$ is generated from the GRU cell state $\bfc = (c_1, \dots, c_n)$ via the following mechanism
\begin{align}
y_i\dti \;=\; c_i\dti \, \fun{H}{ c_i\dti - \vartheta_i } \,, \quad \fun{H}{x} = \begin{cases} 1, \quad x \geq 0 \\ 0, \quad x<0 \end{cases}
\end{align}
Only the sparse output $\bfy$ is communicated between neurons to compute the update gate $\bfu$ and the reset gate $\bfr$ of the GRU
\begin{align}
    \bfu\dti &= \fun{\sigma}{ \bfW_{u}  \concat{ \bfx\dti,\; \bfy\dtim1 } + \bfb_u } \\
    \bfr\dti &= \fun{\sigma}{\bfW_{r} \concat{ \bfx\dti,\; \bfy\dtim1 } + \bfb_r } \, .
\end{align}
As outlined in \cite{mukherji2023activity}, the sparse state $\bfy$ and the gates $\bfu$ and $\bfr$ compute a proposed state $\bfz$ and the new cell state $\bfc$
\begin{align}
    \bfz\dti &= \fun{g}{ \bfW_z \concat{ \bfx\dti,\;  \bfr\dti \hp\; \bfy\dtim1 } + \bfb_z } \\
    \bfc\dti &= \bfu\dti \hp\; \bfz\dti  + (1-\bfu\dti) \hp\; \bfc\dtim1 - \bfs\dti\, .
\end{align}
Similar to biologically plausible spiking neural networks, \cite{subramoney2023efficient} subtract already communicated signals $\bfy$ from the cell state via the reset term 
\(\bfs\dti = \bfth \fun{H}{\bfc\dti - \bfth}\). 
During training, the surrogate function 
\(
    \frac{\mathrm{d}H}{\mathrm{d}c} = \lambda ~ \max
    \left(1 - \lvert c \rvert / \epsilon\right)
\)
provides gradients below the threshold. 

\subsection{Language Modeling with EGRU}
Word-level language modeling is a popular benchmark task to measure the performance of sequence models, including RNNs.
A language model processes a sequence of words $w_1, \dots, w_t\in\mathcal{D}$ from a dictionary $\mathcal{D}$, 
and predicts the conditional distribution $p(w_{t+1}|w_1, \dots, w_t)$.
Its training objective is minimizing the cross entropy $H(p, q)$ between this prediction $p$ and a one-hot encoding $q$ of the actual next word in the sequence.
The standard metric for measuring performance is perplexity (PPL), the exponential cross entropy $e^{H(p, q)}$.
Artificial texts can be generated by a trained language model by iterative sampling from the next-word distribution $p$ predicted by the model.

We trained three EGRU layers without skip connections to processes word embedding vectors drawn from a learned look-up table similar to \cite{Merity2018}.
The model estimates the likelihood of the next word in a sequence by computing the dot-product similarity between the output vector of the final EGRU layer and all word embedding vectors of the dictionary.
Softmax applied to this set of values serves as an estimate of the conditional distribution $p$.
The dimension of word embedding vectors and the final layer cell state was \num{750}.
The dimension of intermediate layers' cell state was \num{1350}.
We used a model from~ Mukherji et al. ~\cite{mukherji2023activity} trained on the WikiText-2 dataset \cite{merity2017pointer} with a parameter sparsity of \SI{95}{\percent} per weight matrix.
The model weights were stored on SRAM in a Sparse CSR format.
The three EGRU layers were implemented on 150 PEs.

\subsection{DVS gesture recognition}

We also evaluated our model on gesture prediction, using the DVS128 Gesture Dataset \cite{amirDVSdataset}.
This dataset contains 11 gestures from 29 subjects recorded with a DVS128 event camera \cite{lichtsteiner_128_2006}. 
Each event encodes a relative change of illumination and is given as spatio-temporal coordinates of X/Y position on the 128×128-pixel sensor and time stamp.

Our model consisted of a CNN feature extraction head and 2 EGRU layers of 256 units each.
The dimension of features extracted from the CNN was \num{512}.
Finally a linear layer was used to predict the class of the gesture.
For this task we used only dense weights and stored them directly on SRAM since the model is small enough to fit in local memory.
The two EGRU layers were implemented on 128 PEs.

\section{SpiNNaker2 implementation}

\subsection{Implementation of EGRU on a single processing element}
We were able to fit the simplest EGRU model on a single PE of SpiNNaker2. 
There are three operations that need to performed as part of EGRU algorithm:
1) input matrix multiplication
2) recurrent matrix multiplication and 
3) point-wise operations.
For a single PE implementation, we can simply execute these operations sequentially.
There is no data transfer needed as all the results are available in local memory.
Although there is a Multiply-accumulate (MAC) accelerator on SpiNNaker2, we do not use it in this application to take full advantage of EGRU's dynamic sparsity.

\subsection{Parallelization approaches}

\begin{figure*}[htb]
    \centering
    \includegraphics[width=0.8\linewidth]{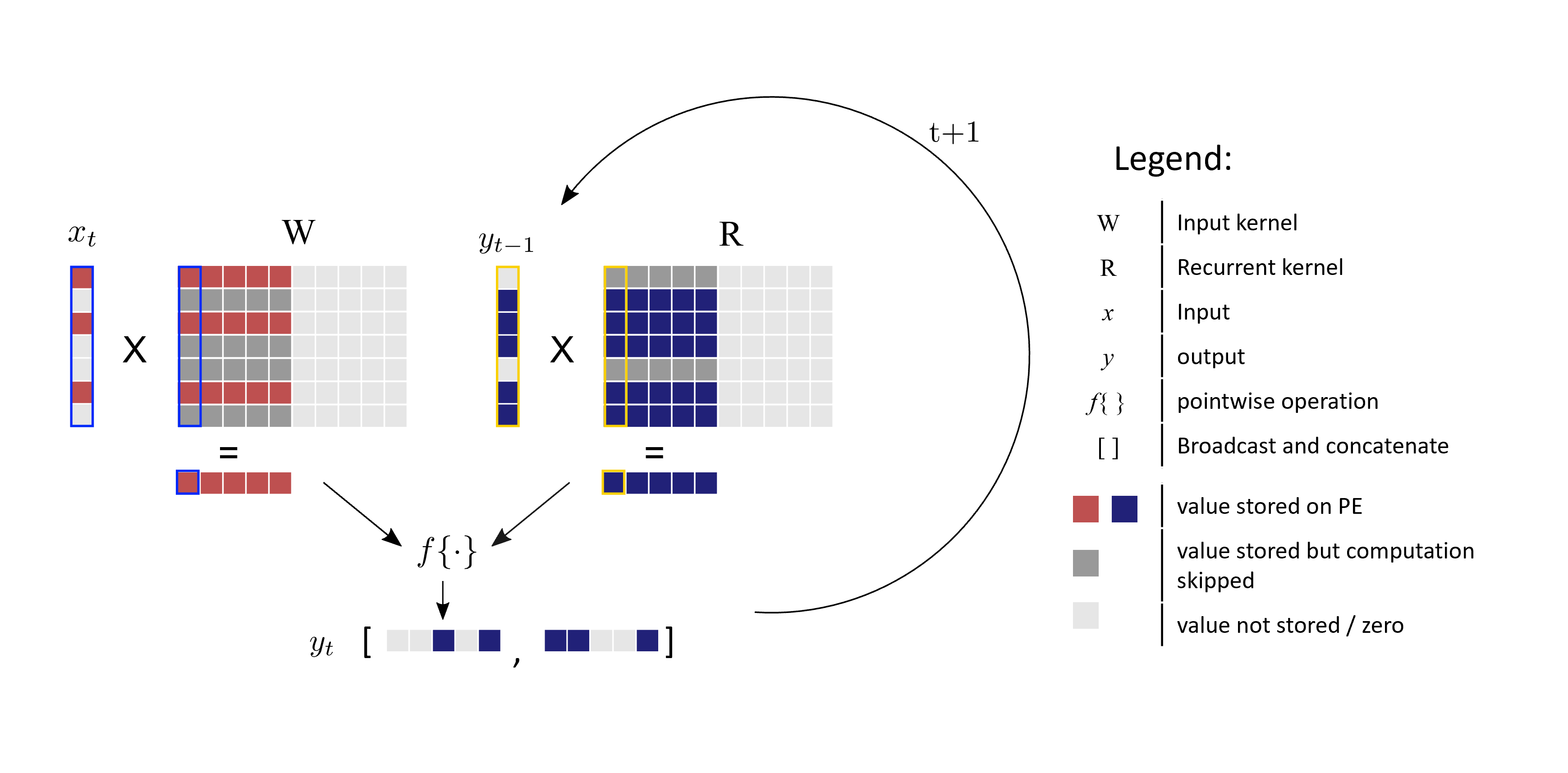}
    \caption{EGRU operations and distribution strategy: This figure shows computation performed on a single PE as part of a multi-core implementation, the grayed out portions are computed on other PEs}
    \label{fig:egru-ops}
\end{figure*}

Since any realistic model, including our larger EGRU models, will not be small enough to fit onto a single PE, we need to split the network over multiple PEs.
To do this, we split the network and place the neurons on different PEs.
This approach reduces the communication and synchronization required within the network.
The output generated by the units placed on a single PE determined the output of that PE.
This output $\bfy\dti$, at time $t$, needed to be broadcast to the rest of the units in the EGRU layer.
On receiving such a broadcast each PE concatenated the outputs from all other PEs together with the output of the units stored locally to form the next recurrent input.
This broadcast was implemented by sending internal NoC packets between PEs.
This operation is demonstrated in Fig.~ \ref{fig:egru-ops} and the algorithm is presented in Algorithm.~\ref{alg:multi-pe}

With this parallelization, we only split one dimension of the $R$ weight matrix.
Since the second dimension was as large as the number of units in a layer, the recurrent weight matrix was still too large to fit in individual PE memory.
To mitigate this, we used a \num{95}\% pruned EGRU model for language modeling.
The pruned weights were stored in compressed sparse row (CSR) format.
In this format the non-zero (NZ) elements of the matrix were represented using three one-dimensional arrays.
These contained NZ values, column indices of the NZ elements and the extents of rows, which required  $2*NZ + N\_rows + 1$ memory.

\begin{algorithm}
\begin{algorithmic}
\Procedure{EGRU}{}
  \State \textbf{Input:}
    \State - Network Configuration Parameters
    \State - Input Data
    \State - Output Data Destination
  \State \textbf{Output:} Processed output data

  \State \textbf{Initialization:}
    \State - Initialize temporary variables.

  \While{run is true}
    \State - Check for input data availability.
    \State - Process input data and prepare it for computation.

    \For{each time step $t$}
      \State - $Wx \gets $   Matrix Multiplication: $W_x\times x_t$.
      \State - $Rh \gets $  Matrix Multiplication: $W_r\times y_{t-1}$.
      \State - Point-wise operation on $Wx$ and $Rh$.
      \State - Store output data.
    \EndFor

    \State - Wait for host to read output data.

    \If{run is false}
      \State - Stop.
    \EndIf
    
  \EndWhile

\EndProcedure
\caption{EGRU algorithm for multi PE implementation}
\label{alg:multi-pe}
\end{algorithmic}
\end{algorithm}

\subsection{Dataset and pre-processing}
\subsubsection{Language Modeling}
The model was trained and validated on WikiText-2 dataset.
The text was tokenised and split into sequence of length 70.
The embeddings were pre-computed and transferred to the LPDDR4 memory.

\subsubsection{DVS}

We combined the DVS raw event times into `frames' by binning them over time windows of 25 ms, and
then downscaled them to 32$\times$32 pixels using a maxpool layer.
The dataset was pre-processed and the features extracted using the CNN head.
The extracted features were stored in the LPDDR4 memory.

\section{Results}
\begin{figure}[htbp]
\centerline{\includegraphics[width=0.6\linewidth]{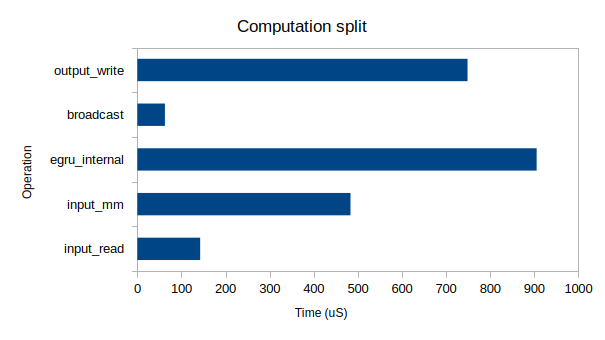}}
\caption{Profiling of internal EGRU operation for Language modeling. EGRU model has 1350 units, 3 layers and 95\% pruned. Weights stored in sparse CSR format.}

\label{fig:lm_compute_timing}
\end{figure}

We measured the time required by the EGRU operations using an internal timer.
This timer ticks at 1 MHz rate and decrements a counter.
We logged the timer value at various points in the algorithm to estimate the time spent by the algorithm at each stage.
The results of this profiling are shown in Fig.~\ref{fig:lm_compute_timing}.
As can be seen, the most expensive part was the recurrent matrix multiplication (\textit{egru\_internal}). 
Broadcasting of layer activations was a comparably cheap operation, since the broadcast uses efficient NoC packets to communicate. 
The bottleneck of the algorithm was therefore found to be dominated by memory reading and writing rather than communication, for the single chip case. 
However, this might not be the case for multi-chip communication.

\begin{table}[htbp]
\caption{Memory footprint of 95\% Pruned EGRU model. Weight matrices stored in sparse CSR format. Total available instruction memory is 32 KB and the available Data memory is 96 KB.}
\centering
\begin{tabular}{@{}lllll@{}}
\toprule
            & Instruction & Debug & Weights & Variables\\ 
\midrule
Memory (KB) & 17.9  & 5    & 88.3  & 2.6              \\ 
\bottomrule
\end{tabular}

\label{tab:csr-mem-footprint}
\end{table}

\begin{table}[htbp]
\caption{EGRU LM on SpiNNaker2 comparison with GPU}
\centering
\begin{tabular}{@{}lll@{}}
\toprule
Measurement & Nvidia A100 & SpiNNaker2 \\ \midrule
Batch size  & 1           & 1          \\
Power (W)   & 60         & 0.39       \\
Time (mS)   & 19.9          & 170.25     \\
Energy (J)  &  1.1935           & 0.0653    \\ 
Test PPL    & 81.4        & 81.4       \\ \bottomrule
\end{tabular}

\label{tab:lm_gpu_compare}
\end{table}

\begin{table}[htbp]
\caption{EGRU DVS gesture prediction on SpiNNaker2 comparison with GPU. Time and Energy measurements normalized over batch size.}
\centering
\begin{tabular}{@{}lllll@{}}
\toprule
Measurement & \multicolumn{2}{c}{Nvidia A100} & \multicolumn{2}{c}{SpiNNaker2}  \\ \midrule
Batch size  & 1   &    6    & 1    & 6      \\
Power (W)   & 61.0  &   61.0    & 0.39 & 0.42      \\
Time (mS)   &  1.09  &  0.19   & 60.19  & 56.20  \\
Energy (J)  &  0.067  &  0.011  & 0.023  & 0.023  \\
Accuracy (\%) & 96.83 & 96.83 & 96.83 & 96.83 \\ \bottomrule
\end{tabular}
\label{tab:dvs_gpu_compare}
\end{table}

\subsection{Power and energy consumption}

The power consumption of the EGRU language model is shown in Table.~\ref{tab:lm_gpu_compare}.
We show that for inference, our implementation on SpiNNaker2 consumes only a fraction of a Watt.
Whereas the time required on SpiNNaker2 scales linearly with batch size, the GPU can process larger batch sizes in the same time.
Hence at larger batch sizes GPU tend to be more efficient.
This is only shown for the DVS task because the LM task has a memory bottleneck that only allows a batch size of one on a single chip.
See Table.~\ref{tab:dvs_gpu_compare} for this energy comparison.
The test accuracy on the classification task was identical on both GPU and SpiNNaker2 implementation, demonstrating numerical equivalence, since we perform 32-bit floating point operations on both architectures.

\section{Outlook}

The successful implementation of a language model on the SpiNNaker2 chip using the EGRU event-based architecture represents a significant milestone in the field of neuromorphic computing.
We compare our implementation with the one on the Nvidia GPU and show a real energy advantage in the single batch size setting, which we expect to be the most relevant for inference, especially on edge devices.
We also identified several bottlenecks in our implementation that need improvement for even further efficiency.
In particular, quantizing the model will allow us to work with even tighter memory constraints to fit larger models and further increase energy efficiency, albeit with a slight reduction in task performance.
Quantization will also allow us to take advantage of the MAC accelerator available on the SpiNNaker2 chip.
We also plan to further scale the deployed model to multiple chips, as SpiNNaker2 is designed for efficient distributed computing.

Scaling up neuromorphic language models to more contemporary large sizes by harnessing very recent innovations in recurrent architectures~\cite{gu2023mamba, poli2023hyena} is the self-evident next step.
However, bringing the performance on par with standard deep learning also suggests expanding the range of real-world applications for neuromorphic hardware in future work, including real-time applications, which they are well suited for.
Overall, our implementation has demonstrated, for the first time, that challenging machine learning tasks are not beyond the scope of neuromorphic computing and heralds the beginning of more mainstream use of neuromorphic devices as complementary to GPUs for appropriate use cases.

\section*{Acknowledgment}

This work was partially funded by the German Federal Ministry of Education and Research (BMBF) and the free state of Saxony within the ScaDS.AI center of excellence for AI research.
This work was partially funded by the German BMBF within the KI-ASIC project (16ES0996).
Khaleel Khan is Funded by the German Federal Ministry of Education and Research (BMBF), funding reference 16ME0729K, joint project "EVENTS".
Mark Schöne is fully funded by the Bosch Research Foundation.
David Kappel is funded by the German Federal Ministry for Economic Affairs and Climate Action (BMWK) project ESCADE (01MN23004A).
Christian Mayr is funded by the German Research Foundation (DFG, Deutsche Forschungsgemeinschaft) as part of Germany’s Excellence Strategy – EXC 2050/1 – Project ID 390696704 – Cluster of Excellence “Centre for Tactile Internet with Human-in-the-Loop” (CeTI) of Technische Universität Dresden.

\clearpage
\bibliographystyle{unsrt}
\bibliography{references}

\begin{thebibliography}{10}

\bibitem{Barham2019}
Paul Barham and Michael Isard.
\newblock Machine learning systems are stuck in a rut.
\newblock In {\em Proceedings of the Workshop on Hot Topics in Operating
  Systems}, HotOS '19, page 177–183, New York, NY, USA, 2019. Association for
  Computing Machinery.

\bibitem{frankle2018lottery}
Jonathan Frankle and Michael Carbin.
\newblock The lottery ticket hypothesis: Finding sparse, trainable neural
  networks.
\newblock {\em arXiv preprint arXiv:1803.03635}, 2018.

\bibitem{Hoefler2021}
Torsten Hoefler, Dan Alistarh, Tal Ben-Nun, Nikoli Dryden, and Alexandra Peste.
\newblock Sparsity in deep learning: Pruning and growth for efficient inference
  and training in neural networks.
\newblock {\em J. Mach. Learn. Res.}, 22(1), jan 2021.

\bibitem{Neil2017}
Daniel Neil, Jun~Haeng Lee, Tobi Delbruck, and Shih-Chii Liu.
\newblock Delta networks for optimized recurrent network computation.
\newblock In Doina Precup and Yee~Whye Teh, editors, {\em Proceedings of the
  34th International Conference on Machine Learning}, volume~70 of {\em
  Proceedings of Machine Learning Research}, pages 2584--2593. PMLR, 06--11 Aug
  2017.

\bibitem{li2023}
Zonglin Li, Chong You, Srinadh Bhojanapalli, Daliang Li, Ankit~Singh Rawat,
  Sashank~J. Reddi, Ke~Ye, Felix Chern, Felix Yu, Ruiqi Guo, and Sanjiv Kumar.
\newblock The lazy neuron phenomenon: On emergence of activation sparsity in
  transformers.
\newblock In {\em The Eleventh International Conference on Learning
  Representations}, 2023.

\bibitem{subramoney2023efficient}
Anand Subramoney, Khaleelulla~Khan Nazeer, Mark Sch{\"o}ne, Christian Mayr, and
  David Kappel.
\newblock Efficient recurrent architectures through activity sparsity and
  sparse back-propagation through time.
\newblock In {\em The Eleventh International Conference on Learning
  Representations}, 2023.

\bibitem{relu}
Iman Mirzadeh, Keivan Alizadeh, Sachin Mehta, Carlo C~Del Mundo, Oncel Tuzel,
  Golnoosh Samei, Mohammad Rastegari, and Mehrdad Farajtabar.
\newblock Relu strikes back: Exploiting activation sparsity in large language
  models.
\newblock In {\em NeurIPS}, 2023.

\bibitem{Horowitz2014}
Mark Horowitz.
\newblock 1.1 computing's energy problem (and what we can do about it).
\newblock In {\em 2014 IEEE International Solid-State Circuits Conference
  Digest of Technical Papers (ISSCC)}, pages 10--14, 2014.

\bibitem{gonzalez2023spinnaker}
Hector~Andres Gonzalez, Jiaxin Huang, Florian Kelber, Khaleelulla~Khan Nazeer,
  Tim~Hauke Langer, Chen Liu, Matthias~Aleander Lohrmann, Amirhossein Rostami,
  Mark Sch{\"o}ne, Bernhard Vogginger, Timo Wunderlich, Yexin Yan, Mahmoud Akl,
  and Christian Mayr.
\newblock Spi{NN}aker2: A large-scale neuromorphic system for event-based and
  asynchronous machine learning.
\newblock In {\em First Workshop on Machine Learning with New Compute
  Paradigms}, 2023.

\bibitem{Vaswani2017}
Ashish Vaswani, Noam Shazeer, Niki Parmar, Jakob Uszkoreit, Llion Jones,
  Aidan~N. Gomez, \L{}ukasz Kaiser, and Illia Polosukhin.
\newblock Attention is all you need.
\newblock In {\em Proceedings of the 31st International Conference on Neural
  Information Processing Systems}, NIPS'17, page 6000–6010, Red Hook, NY,
  USA, 2017. Curran Associates Inc.

\bibitem{Gu2022}
Albert Gu, Karan Goel, and Christopher Re.
\newblock Efficiently modeling long sequences with structured state spaces.
\newblock In {\em International Conference on Learning Representations}, 2022.

\bibitem{Peng2023}
Bo~Peng, Eric Alcaide, Quentin Anthony, Alon Albalak, Samuel Arcadinho, Huanqi
  Cao, Xin Cheng, Michael Chung, Matteo Grella, Kranthi~Kiran GV, Xuzheng He,
  Haowen Hou, Przemyslaw Kazienko, Jan Kocon, Jiaming Kong, Bartlomiej Koptyra,
  Hayden Lau, Krishna Sri~Ipsit Mantri, Ferdinand Mom, Atsushi Saito, Xiangru
  Tang, Bolun Wang, Johan~S. Wind, Stansilaw Wozniak, Ruichong Zhang, Zhenyuan
  Zhang, Qihang Zhao, Peng Zhou, Jian Zhu, and Rui-Jie Zhu.
\newblock Rwkv: Reinventing rnns for the transformer era, 2023.

\bibitem{freund2020graphcore}
Karl Freund and Patrick Moorhead.
\newblock The graphcore second-generation ipu, 2020.

\bibitem{lie2023cerebras}
Sean Lie.
\newblock Cerebras architecture deep dive: First look inside the
  hardware/software co-design for deep learning.
\newblock {\em IEEE Micro}, 43(3):18--30, 2023.

\bibitem{kim2023full}
Sehoon Kim, Coleman Hooper, Thanakul Wattanawong, Minwoo Kang, Ruohan Yan,
  Hasan Genc, Grace Dinh, Qijing Huang, Kurt Keutzer, Michael~W Mahoney, et~al.
\newblock Full stack optimization of transformer inference: a survey.
\newblock {\em arXiv preprint arXiv:2302.14017}, 2023.

\bibitem{keller202395}
Ben Keller, Rangharajan Venkatesan, Steve Dai, Stephen~G Tell, Brian Zimmer,
  Charbel Sakr, William~J Dally, C~Thomas Gray, and Brucek Khailany.
\newblock A 95.6-tops/w deep learning inference accelerator with per-vector
  scaled 4-bit quantization in 5 nm.
\newblock {\em IEEE Journal of Solid-State Circuits}, 58(4):1129--1141, 2023.

\bibitem{zhang2022spike}
Jiyuan Zhang, Lulu Tang, Zhaofei Yu, Jiwen Lu, and Tiejun Huang.
\newblock Spike transformer: Monocular depth estimation for spiking camera.
\newblock In {\em European Conference on Computer Vision}, pages 34--52.
  Springer, 2022.

\bibitem{zhu2023spikegpt}
Rui-Jie Zhu, Qihang Zhao, and Jason~K Eshraghian.
\newblock Spikegpt: Generative pre-trained language model with spiking neural
  networks.
\newblock {\em arXiv preprint arXiv:2302.13939}, 2023.

\bibitem{bal2023spikingbert}
Malyaban Bal and Abhronil Sengupta.
\newblock Spikingbert: Distilling bert to train spiking language models using
  implicit differentiation.
\newblock {\em arXiv preprint arXiv:2308.10873}, 2023.

\bibitem{sun2018fpga}
Zhanrui Sun, Yongxin Zhu, Yu~Zheng, Hao Wu, Zihao Cao, Peng Xiong, Junjie Hou,
  Tian Huang, and Zhiqiang Que.
\newblock Fpga acceleration of lstm based on data for test flight.
\newblock In {\em 2018 IEEE International Conference on Smart Cloud
  (SmartCloud)}, pages 1--6. IEEE, 2018.

\bibitem{conti2018chipmunk}
Francesco Conti, Lukas Cavigelli, Gianna Paulin, Igor Susmelj, and Luca Benini.
\newblock Chipmunk: A systolically scalable 0.9 mm 2, 3.08 gop/s/mw@ 1.2 mw
  accelerator for near-sensor recurrent neural network inference.
\newblock In {\em 2018 IEEE Custom Integrated Circuits Conference (CICC)},
  pages 1--4. IEEE, 2018.

\bibitem{smagulova2019survey}
Kamilya Smagulova and Alex~Pappachen James.
\newblock A survey on lstm memristive neural network architectures and
  applications.
\newblock {\em The European Physical Journal Special Topics},
  228(10):2313--2324, 2019.

\bibitem{rao2022long}
Arjun Rao, Philipp Plank, Andreas Wild, and Wolfgang Maass.
\newblock A long short-term memory for ai applications in spike-based
  neuromorphic hardware.
\newblock {\em Nature Machine Intelligence}, 4(5):467--479, 2022.

\bibitem{Ali2020}
Ali Lotfi~Rezaabad and Sriram Vishwanath.
\newblock Long short-term memory spiking networks and their applications.
\newblock In {\em International Conference on Neuromorphic Systems 2020}, ICONS
  2020, New York, NY, USA, 2020. Association for Computing Machinery.

\bibitem{SigmaDelta2017}
Peter O'Connor and Max Welling.
\newblock Sigma delta quantized networks.
\newblock In {\em International Conference on Learning Representations}, 2017.

\bibitem{Spartus2022}
Chang Gao, Tobi Delbruck, and Shih-Chii Liu.
\newblock Spartus: A 9.4 top/s fpga-based lstm accelerator exploiting
  spatio-temporal sparsity.
\newblock {\em IEEE Transactions on Neural Networks and Learning Systems},
  pages 1--15, 2022.

\bibitem{Cho2014}
Kyunghyun Cho, Bart van Merrienboer, {\c{C}}aglar G{\"{u}}l{\c{c}}ehre, Dzmitry
  Bahdanau, Fethi Bougares, Holger Schwenk, and Yoshua Bengio.
\newblock Learning phrase representations using {RNN} encoder-decoder for
  statistical machine translation.
\newblock In Alessandro Moschitti, Bo~Pang, and Walter Daelemans, editors, {\em
  Proceedings of the 2014 Conference on Empirical Methods in Natural Language
  Processing, {EMNLP} 2014, October 25-29, 2014, Doha, Qatar, {A} meeting of
  SIGDAT, a Special Interest Group of the {ACL}}, pages 1724--1734. {ACL},
  2014.

\bibitem{mukherji2023activity}
Rishav Mukherji, Mark Sch{\"o}ne, Khaleelulla~Khan Nazeer, Christian Mayr, and
  Anand Subramoney.
\newblock Activity sparsity complements weight sparsity for efficient {RNN}
  inference.
\newblock In {\em First Workshop on Machine Learning with New Compute
  Paradigms}, 2023.

\bibitem{Merity2018}
Stephen Merity, Nitish~Shirish Keskar, and Richard Socher.
\newblock Regularizing and optimizing {LSTM} language models.
\newblock In {\em International Conference on Learning Representations}, 2018.

\bibitem{merity2017pointer}
Stephen Merity, Caiming Xiong, James Bradbury, and Richard Socher.
\newblock Pointer sentinel mixture models.
\newblock In {\em International Conference on Learning Representations}, 2017.

\bibitem{amirDVSdataset}
Arnon Amir, Brian Taba, David Berg, Timothy Melano, Jeffrey {McKinstry},
  Carmelo~Di Nolfo, Tapan Nayak, Alexander Andreopoulos, Guillaume Garreau,
  Marcela Mendoza, Jeff Kusnitz, Michael Debole, Steve Esser, Tobi Delbruck,
  Myron Flickner, and Dharmendra Modha.
\newblock A low power, fully event-based gesture recognition system.
\newblock page~10.

\bibitem{lichtsteiner_128_2006}
P.~Lichtsteiner, C.~Posch, and T.~Delbruck.
\newblock A 128 x 128 120db 30mw asynchronous vision sensor that responds to
  relative intensity change.
\newblock In {\em 2006 {IEEE} International Solid State Circuits Conference -
  Digest of Technical Papers}, pages 2060--2069.
\newblock {ISSN}: 2376-8606.

\bibitem{gu2023mamba}
Albert Gu and Tri Dao.
\newblock Mamba: {{Linear-Time Sequence Modeling}} with {{Selective State
  Spaces}}, December 2023.

\bibitem{poli2023hyena}
Michael Poli, Stefano Massaroli, Eric Nguyen, Daniel~Y. Fu, Tri Dao, Stephen
  Baccus, Yoshua Bengio, Stefano Ermon, and Christopher R{\'e}.
\newblock Hyena {{Hierarchy}}: {{Towards Larger Convolutional Language
  Models}}, April 2023.

\end{thebibliography}

\end{document}